\documentclass[11pt,a4paper]{article}

\usepackage[margin=1in]{geometry}
\usepackage{graphicx}
\usepackage{booktabs}
\usepackage{amsmath}
\usepackage{amssymb}
\usepackage{url}
\usepackage{hyperref}
\usepackage{xcolor}
\usepackage{authblk}
\usepackage{caption}
\usepackage{microtype}

\hypersetup{colorlinks=true,linkcolor=blue,urlcolor=blue,citecolor=blue}
\graphicspath{{figures/}}

\title{Domain Fine-Tuning vs.\ Retrieval-Augmented Generation\\
       for Medical Multiple-Choice Question Answering:\\
       A Controlled Comparison at the 4B-Parameter Scale}

       \author{
        Avi-ad Avraam Buskila\\
        \small Department of Information Science and Applied Artificial Intelligence, Bar-Ilan University, Ramat-Gan, Israel\\
        \small \texttt{[aviad-avraam.buskila@biu.ac.il]}
      }

\date{April 2026}

\begin{document}

\maketitle

\begin{abstract}
Practitioners deploying small open-weight large language models (LLMs) for
medical question answering face a recurring design choice: invest in a
domain-fine-tuned model, or keep a general-purpose model and inject
domain knowledge at inference time via retrieval-augmented generation
(RAG). We isolate this trade-off by holding model size, prompt template,
decoding temperature, retrieval pipeline, and evaluation protocol fixed,
and varying only \emph{(i)} whether the model has been domain-adapted
(\textsc{Gemma~3 4B} vs.\ \textsc{MedGemma 4B}, both 4-bit quantized and
served via Ollama) and \emph{(ii)} whether retrieved passages from a
medical knowledge corpus are inserted into the prompt. We evaluate all
four cells of this $2\!\times\!2$ design on the full MedQA-USMLE
4-option test split (1{,}273 questions) with three repetitions per
question (15{,}276 LLM calls). Domain fine-tuning yields a $+6.8$
percentage-point gain in majority-vote accuracy over the general 4B
baseline (53.3\% vs.\ 46.4\%, McNemar $p<10^{-4}$). RAG over MedMCQA
explanations does not produce a statistically significant gain in
either model, and in the domain-tuned model the point estimate is
slightly negative ($-1.9$\,pp, $p=0.16$). At this scale and on this
benchmark, domain knowledge encoded in weights dominates domain
knowledge supplied in context. We release the full experiment code and
JSONL traces to support replication.\footnote{Release v1:
\url{https://github.com/aviad-buskila/what_makes_it_right/tree/v1}.}
\end{abstract}

\section{Introduction}

Building reliable medical question-answering systems with small,
locally-runnable language models is attractive for cost, latency, and
privacy reasons, but small models lag larger frontier models on
clinical benchmarks~\cite{singhal2023medpalm,nori2023gpt4medical}. Two
practical interventions are available to a practitioner who is
constrained to a fixed parameter budget:
\begin{enumerate}
  \item \textbf{Fine-tuning / continued pretraining} on medical
    corpora, which encodes domain knowledge into the model's weights
    (e.g., \textsc{MedGemma}~\cite{sellergren2025medgemma} on the
    \textsc{Gemma}~\cite{team2024gemma} backbone).
  \item \textbf{Retrieval-augmented generation
    (RAG)}~\cite{lewis2020rag}, which leaves the weights untouched and
    instead conditions the model on retrieved passages from a domain
    corpus at inference time.
\end{enumerate}

Both interventions have well-documented benefits on medical
benchmarks~\cite{xiong2024benchmarkingrag,wang2024augmentingblackbox},
but they are usually compared in different studies, with different
backbones, different evaluation sets, and different retrieval
pipelines. The practitioner's actual question --- \emph{at a fixed
parameter budget, which intervention buys more accuracy?} --- is
rarely answered head-to-head.

This paper conducts that head-to-head comparison in a controlled
$2\!\times\!2$ design (general vs.\ domain-tuned model)~$\times$~(no
RAG vs.\ RAG) at the 4B-parameter scale. We deliberately use small
quantized models served via Ollama because that is the deployment
profile most accessible to clinicians, students, and small clinics.

\paragraph{Contributions.}
\begin{itemize}
  \item A controlled, paired comparison of domain fine-tuning and RAG
    on the full MedQA-USMLE 4-option test split (1{,}273 questions, 3
    repetitions, 15{,}276 LLM calls).
  \item Statistical evidence (McNemar's test on paired predictions)
    that, at 4B parameters, domain fine-tuning gives a significant
    accuracy gain whereas a competitive RAG pipeline does not.
  \item A fully reproducible open-source pipeline (Ollama + ChromaDB +
    \texttt{nomic-embed-text}~\cite{nomic2024embed,chromadb,ollama2024})
    with deterministic temperature and per-question repetitions.
\end{itemize}

\section{Related Work}

\paragraph{Medical LLMs.} Singhal et al.\ established that scaling
plus instruction tuning yields strong USMLE-style
performance~\cite{singhal2023medpalm}, and \textsc{GPT-4} demonstrated
near- or above-passing performance on USMLE-style
items~\cite{nori2023gpt4medical}. \textsc{MedGemma} continues this
trend in the open-weight regime by adapting \textsc{Gemma~3} to medical
text and images~\cite{sellergren2025medgemma}.

\paragraph{Retrieval-augmented generation for medicine.}
\textsc{MIRAGE}~\cite{xiong2024benchmarkingrag} and concurrent
work~\cite{wang2024augmentingblackbox} show RAG can lift accuracy on
medical QA, especially for general-purpose backbones. The reported
gains are typically larger when the retrieval corpus closely matches
the question source (e.g., textbook-grounded questions retrieved over
the same textbooks).

\paragraph{Direct comparison.} Few studies hold the backbone, prompt,
and evaluation set fixed while toggling \emph{both} fine-tuning and
RAG. Our work closes that gap at a small, deployment-relevant scale.

\section{Method}

\subsection{Design}

We use a fully crossed $2\!\times\!2$ design over two binary factors:
\textsc{Domain-Tuned} (yes / no) and \textsc{RAG} (yes / no). All
other factors --- prompt template, decoding temperature, max retrieval
budget, evaluation set, repetitions, and aggregation --- are held
constant. The four resulting setups are summarized in
Table~\ref{tab:setups}.

\begin{table}[h]
\centering
\small
\begin{tabular}{lll}
\toprule
\textbf{Setup} & \textbf{Backbone} & \textbf{Context} \\
\midrule
\textsc{gemma3-4b} & Gemma~3 4B (general) & question only \\
\textsc{gemma3-4b+RAG} & Gemma~3 4B (general) & question + retrieved passages \\
\textsc{medgemma-4b} & MedGemma 4B (domain) & question only \\
\textsc{medgemma-4b+RAG} & MedGemma 4B (domain) & question + retrieved passages \\
\bottomrule
\end{tabular}
\caption{The four experimental setups. Both backbones are served as
4-bit-quantized GGUF builds via Ollama at temperature 0.1.}
\label{tab:setups}
\end{table}

\subsection{Models}

Both backbones are 4B-parameter, 4-bit-quantized, instruction-tuned
checkpoints served locally via Ollama~\cite{ollama2024}. The general
backbone is \texttt{gemma3:4b}~\cite{team2024gemma}; the domain
backbone is a community 4-bit GGUF build of
\textsc{MedGemma}~4B~\cite{sellergren2025medgemma}
(\texttt{edwardlo12/medgemma-4b-it-q4\_k\_m}). Sampling temperature is
fixed to $0.1$ for all calls.

\subsection{Evaluation set}

We use the test split of MedQA-USMLE 4-option~\cite{jin2021medqa}
(1{,}273 items). Each item is a USMLE Step-style multiple-choice
question with four options and a single gold answer.

\subsection{Retrieval pipeline}

To avoid trivially leaking answers, the retrieval corpus is the
\emph{explanation} field of the MedMCQA training
split~\cite{pal2022medmcqa} --- the questions and answer options of
MedMCQA are never indexed, and MedMCQA is disjoint from MedQA. This
makes the corpus a textbook-style proxy rather than a leakage path.

The corpus is chunked at 512 tokens with 50-token overlap, embedded
with \texttt{nomic-embed-text}~\cite{nomic2024embed}, and stored in a
ChromaDB persistent collection~\cite{chromadb}. At inference time we
form a query from the question text concatenated with all four answer
options, retrieve dense candidates, drop chunks with cosine distance
above $0.3$, and re-rank them with a hybrid score $\alpha \cdot
\text{dense} + (1-\alpha) \cdot \text{lexical}$ ($\alpha=0.6$). The
top $k=3$ chunks are inserted into the RAG prompt. Across all
RAG calls, 96.35\% returned a non-empty context (post distance
cutoff) with a mean of 2.89 retrieved chunks per call.

\subsection{Prompting}

Prompts are paraphrases of one another and both enforce the same
machine-readable answer contract \texttt{\{"answer":"A|B|C|D"\}}. The
RAG prompt prepends the retrieved chunks and explicitly tells the
model that the passages \emph{may or may not} be relevant. Both
prompts use the same medical-expert persona. The exact strings are in
the released code (\texttt{src/llm/prompts.py}).

\subsection{Repetitions and aggregation}

Each (question, setup) pair is evaluated three times. The reported
accuracy is the \emph{majority-vote} answer across the three
repetitions; ties are broken by the first repetition. We also report
\emph{consistency} (the average within-setup agreement across the
three repetitions), which is high in all conditions ($\geq 0.99$,
Table~\ref{tab:consistency}), so the choice of aggregation rule has
negligible impact on the conclusions.

\subsection{Statistical testing}

To compare any two setups we use McNemar's exact paired
test~\cite{dietterich1998mcnemar} on majority-vote correctness over
the same 1{,}273 items. We report the discordant cell counts
($a$-only-correct, $b$-only-correct), the test statistic, the $p$-value,
and the binomial 95\% confidence interval on accuracy.

\section{Results}

\subsection{Headline accuracy}

\begin{figure}[h]
\centering
\includegraphics[width=0.85\linewidth]{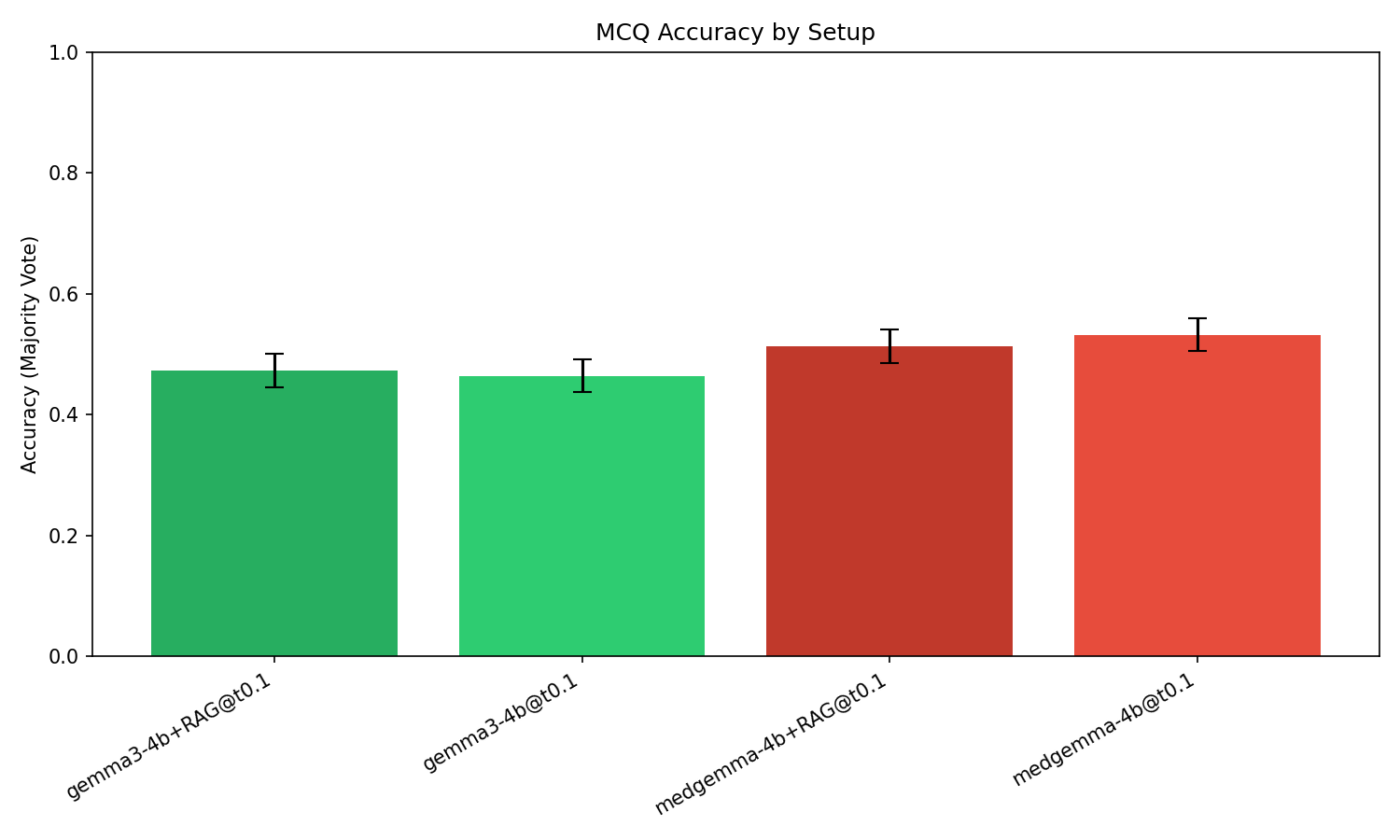}
\caption{Majority-vote accuracy with 95\% confidence intervals. The
$+6.8$\,pp gap from domain fine-tuning is significant; the RAG gap is
not.}
\label{fig:accuracy}
\end{figure}

Majority-vote accuracy and Wilson 95\% confidence intervals for the
four setups appear in Table~\ref{tab:accuracy} and
Figure~\ref{fig:accuracy}. The domain-tuned backbone (\textsc{MedGemma
4B}) leads both with and without RAG, and the best non-domain setup
(\textsc{Gemma~3 4B + RAG}) does not catch up to the worst domain
setup (\textsc{MedGemma 4B + RAG}).

\begin{table}[h]
\centering
\small
\begin{tabular}{lcccc}
\toprule
\textbf{Setup} & \textbf{Accuracy} & \textbf{95\% CI} & \textbf{Correct} & \textbf{N} \\
\midrule
\textsc{gemma3-4b}        & 0.4643 & [0.4370, 0.4917] & 591 & 1273 \\
\textsc{gemma3-4b+RAG}    & 0.4729 & [0.4456, 0.5004] & 602 & 1273 \\
\textsc{medgemma-4b+RAG}  & 0.5137 & [0.4863, 0.5411] & 654 & 1273 \\
\textsc{medgemma-4b}      & \textbf{0.5326} & [0.5051, 0.5599] & \textbf{678} & 1273 \\
\bottomrule
\end{tabular}
\caption{Majority-vote accuracy on the MedQA-USMLE 4-option test split.}
\label{tab:accuracy}
\end{table}

\subsection{Pairwise significance}

\begin{figure}[h]
\centering
\includegraphics[width=0.85\linewidth]{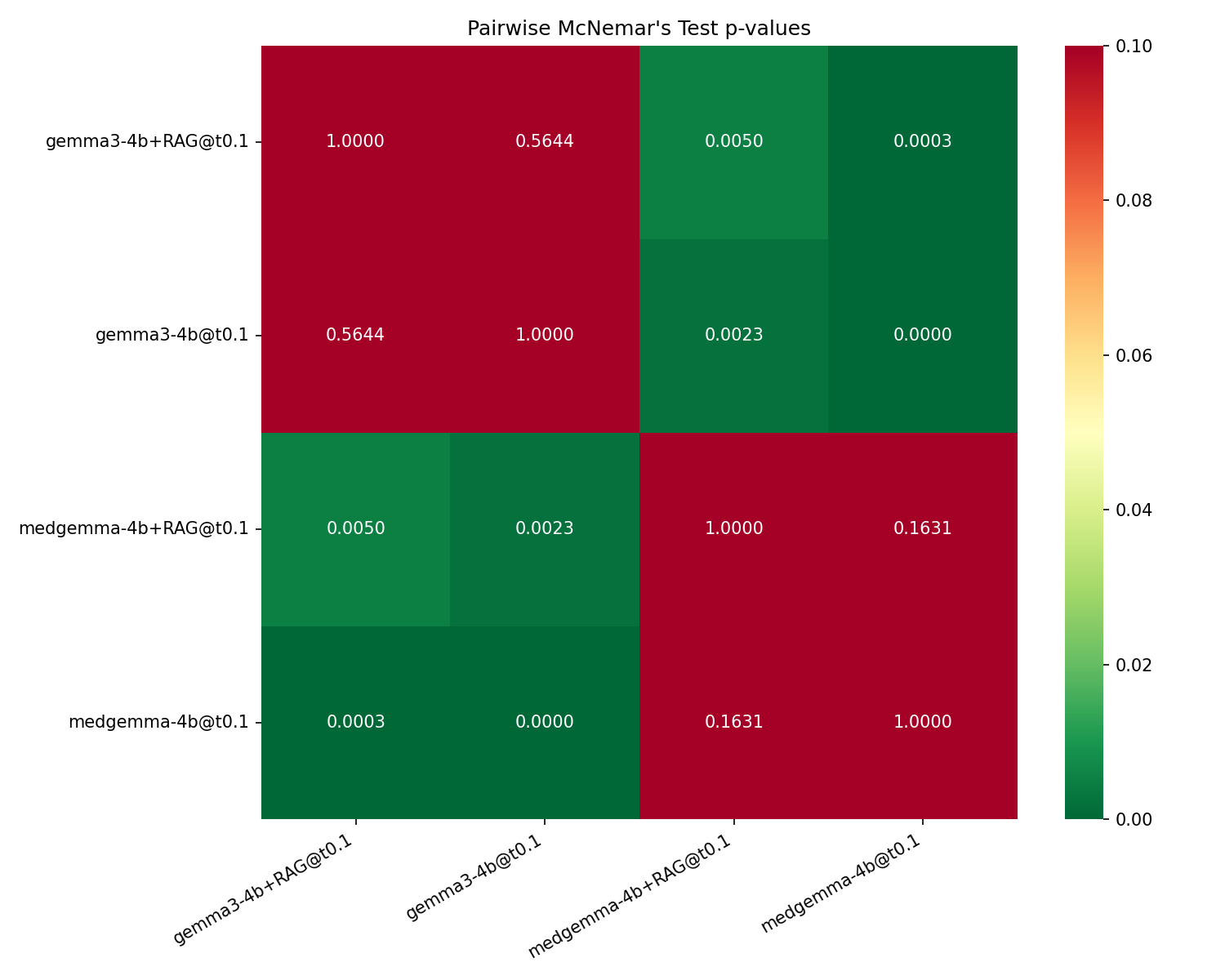}
\caption{Pairwise McNemar $p$-values across the four setups. The two
non-significant cells are precisely the two RAG-toggle comparisons.}
\label{fig:significance}
\end{figure}

McNemar's test on paired predictions
(Table~\ref{tab:mcnemar}, Figure~\ref{fig:significance}) tells a
clear story:
\begin{itemize}
  \item Every comparison that \emph{crosses} the model boundary
    (general vs.\ domain) is significant at $p<0.005$.
  \item Neither comparison \emph{within} a model (base vs.\ +RAG) is
    significant. For \textsc{MedGemma 4B} the point estimate of the
    RAG effect is in fact slightly negative
    ($-1.9$\,pp, $p=0.16$).
\end{itemize}

\begin{table}[h]
\centering
\small
\begin{tabular}{llrrrr}
\toprule
\textbf{Setup A} & \textbf{Setup B} & $a$-only & $b$-only & $\chi^2$ & $p$-value \\
\midrule
\textsc{gemma3-4b+RAG}   & \textsc{gemma3-4b}        & 156 & 145 &  0.33 & 0.5644 \\
\textsc{medgemma-4b+RAG} & \textsc{medgemma-4b}      & 124 & 148 &  1.94 & 0.1631 \\
\textsc{gemma3-4b+RAG}   & \textsc{medgemma-4b+RAG}  & 139 & 191 &  7.88 & 0.0050 \\
\textsc{gemma3-4b}       & \textsc{medgemma-4b}      & 144 & 231 & 19.72 & $<10^{-4}$ \\
\textsc{gemma3-4b+RAG}   & \textsc{medgemma-4b}      & 179 & 255 & 12.96 & 0.0003 \\
\textsc{gemma3-4b}       & \textsc{medgemma-4b+RAG}  & 176 & 239 &  9.26 & 0.0023 \\
\bottomrule
\end{tabular}
\caption{McNemar's test on paired majority-vote correctness.
$a$-only / $b$-only are the discordant cells. Comparisons that cross
the general/domain backbone boundary are significant; comparisons that
toggle RAG within a backbone are not.}
\label{tab:mcnemar}
\end{table}

\subsection{Consistency}

All four setups are highly consistent across repetitions
(Table~\ref{tab:consistency}); consistency is $\geq 0.99$ in every
condition. RAG slightly increases consistency on the general backbone
and slightly decreases it on the domain backbone, but all four values
are within $\sim 0.7$\,pp of each other. The choice of three
repetitions therefore mainly serves to expose decoding noise; it is
not a major source of variance at $T=0.1$.

\begin{table}[h]
\centering
\small
\begin{tabular}{lcc}
\toprule
\textbf{Setup} & \textbf{Consistency} & \textbf{Parse-fail rate} \\
\midrule
\textsc{gemma3-4b}        & 0.9974 & 0.0000 \\
\textsc{gemma3-4b+RAG}    & 0.9992 & 0.0000 \\
\textsc{medgemma-4b}      & 0.9921 & 0.0047 \\
\textsc{medgemma-4b+RAG}  & 0.9919 & 0.0039 \\
\bottomrule
\end{tabular}
\caption{Within-setup agreement across the three repetitions, and
fraction of model outputs that did not match the answer schema after
two retries.}
\label{tab:consistency}
\end{table}

\begin{figure}[h]
\centering
\includegraphics[width=0.7\linewidth]{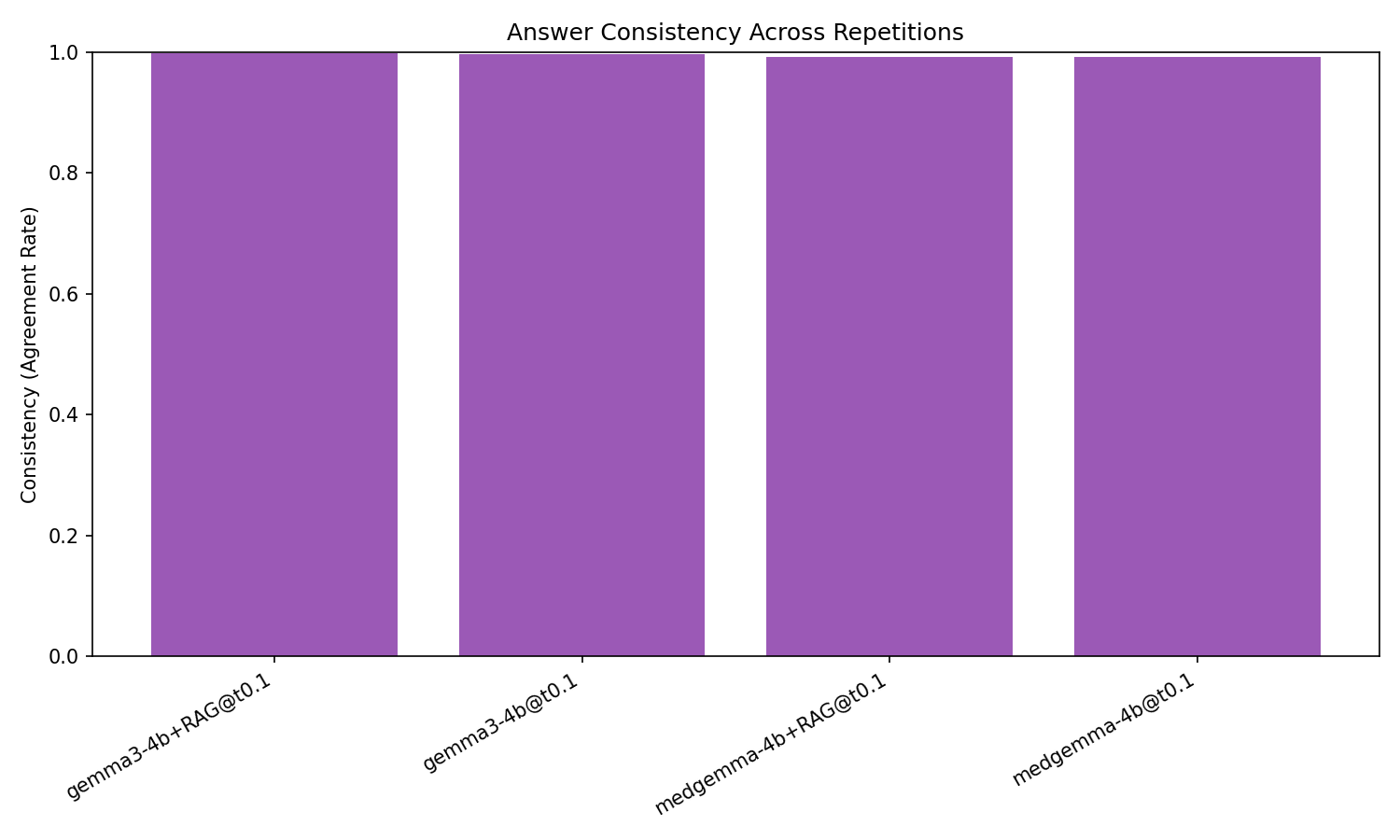}
\caption{Within-setup consistency across repetitions.}
\label{fig:consistency}
\end{figure}

\section{Discussion}

The headline finding is straightforward: for a 4B-parameter model
answering USMLE-style multiple-choice questions, swapping the general
\textsc{Gemma~3} backbone for the domain-adapted \textsc{MedGemma}
backbone delivers a $+6.8$\,pp accuracy gain that is highly
significant on paired data ($p<10^{-4}$), whereas adding a
distance-filtered RAG context retrieved from a textbook-style corpus
does \emph{not} produce a significant gain in either backbone.

\paragraph{Why might RAG fail to help here?}
We see at least four plausible contributors, none of which we can
definitively rule out from this experiment alone. \emph{(i)} USMLE
items are heavily \emph{reasoning}-driven: the diagnosis or next-best
step usually requires integrating findings, not retrieving a fact;
even a perfectly relevant passage may not shortcut the chain of
reasoning. \emph{(ii)} The corpus (MedMCQA explanations) is broad but
not authoritative, and may dilute rather than reinforce the model's
prior. \emph{(iii)} A 4B model has limited capacity to read, ground,
and integrate three new chunks while still solving the question; RAG
gains in the literature are most clearly demonstrated on
larger backbones~\cite{xiong2024benchmarkingrag}. \emph{(iv)}
\textsc{MedGemma} has likely already absorbed much of the textbook
knowledge that the retrieval corpus tries to surface, leaving little
marginal information for RAG to add --- and possibly some interference
when the retrieved passage and the model's prior conflict.

\paragraph{Practical implication.}
For a practitioner choosing how to spend an engineering budget on a
small, locally-deployable medical QA system, our results suggest that
adopting a domain-adapted backbone is a more reliable lever than
building a retrieval pipeline. RAG is not actively harmful here, but
it is not a substitute for in-weights domain knowledge at this scale.

\section{Threats to Validity}

\paragraph{Single benchmark.} We evaluate only on MedQA-USMLE
4-option. Findings may differ on benchmarks where retrieval is more
clearly textbook-grounded (e.g., open-book settings where the answer
appears verbatim in the corpus).

\paragraph{Single corpus and embedding.} We use one retrieval corpus
(MedMCQA explanations) and one embedding model
(\texttt{nomic-embed-text}). A different corpus (e.g., UpToDate,
clinical guidelines, or MedQA-aligned textbooks) or a stronger
embedder could change the RAG result. We mitigated obvious
configuration weakness by tuning \texttt{max\_distance}, $k$, and the
hybrid rerank coefficient on a smaller pilot before locking the
configuration for the full run.

\paragraph{4-bit quantization.} Both backbones are served as 4-bit
GGUF builds. Quantization can shift accuracy by a few points and may
not affect both backbones equally; the absolute numbers should be read
in that light, although the \emph{paired} McNemar comparison still
isolates the effect of the manipulated factor.

\paragraph{Single scale.} Both models are 4B parameters. We cannot
extrapolate to 70B-class or frontier API models, where prior work
suggests RAG gains can be larger.

\paragraph{Possible pretraining contamination.} MedQA is a public
benchmark; either backbone may have seen items during pretraining. We
have no reason to believe contamination differentially favors one
backbone, but we cannot rule out an absolute-level effect.

\section{Conclusion}

In a fully paired $2\!\times\!2$ comparison on the full MedQA-USMLE
test split, domain fine-tuning of a 4B model significantly outperforms
RAG over a textbook-style corpus, while RAG does not significantly
help either the general or the domain-tuned backbone. At the
deployment-relevant scale of small open-weight models served locally,
domain knowledge in the weights appears to dominate domain knowledge
in the prompt. We hope this controlled comparison helps practitioners
prioritize fine-tuning over retrieval pipelines when both options
are available.

\section*{Reproducibility}

Code, configuration files, raw JSONL outputs (15{,}276 records), and
analysis scripts are provided in the project repository, tagged as
release \texttt{v1}:
\url{https://github.com/aviad-buskila/what_makes_it_right/tree/v1}.
The full experiment can be reproduced end-to-end with:
\begin{verbatim}
ollama serve
python scripts/build_rag_index.py --config config.yaml
python scripts/run_experiment.py  --config config.yaml
python scripts/analyze_results.py --experiment medical_mcq_comparison_full_0.1_3r
\end{verbatim}
The exact retrieval, prompt, and decoding hyperparameters are
committed in \texttt{config.yaml}.

\bibliographystyle{plain}
\bibliography{references}

\end{document}